# A Strategy Transfer and Decision Support Approach for Epidemic Control in Experience Shortage Scenarios

Xingyu Xiao[1,2,3], Peng Chen[4,5], Xue Cao [1,2], Kai Liu [1,2], Li Deng[6], Dongjia Zhao[7], Zhe Chen[8], Qing Deng[1,2,#], Feng Yu[9], Hui Zhang[10]

[1]Research Institute of Macro-Safety Science, University of Science and Technology Beijing, Beijing, 100083, China
[2]School of Civil and Resource Engineering, University of Science and Technology Beijing, Beijing, 100083, China
[3]Institute of Nuclear and New Energy Technology, Tsinghua University, Beijing, 100084, China
[4]School of Computer and Communication Engineering, University of Science and Technology Beijing, Beijing, 100083, China
[5] Institute of Software, Chinese Academy of Sciences, Beijing, 100049, China
[6]Central South Architectural Design Institute Co., Wuhan, 430071, China
[7]Xi'an Jiao Tong University, Xi'an, 713599, China
[8]Shanxi University, Taiyuan, 030006, China
[9]Shanghai Jiao Tong University, Shanghai, 200030, China
[10]Deparment of Engineering Physics, Tsinghua University, 100091, Beijing, China
[#]Corresponding author: Qing Deng. Email: dengqing0415@126.com

**Abstract.**
Epidemic outbreaks can cause critical health concerns and severe global economic crises. For countries or regions with new infectious disease outbreaks, it is essential to generate preventive strategies by learning lessons from others with similar risk profiles. A Strategy Transfer and Decision Support Approach (STDSA) is proposed based on the profile similarity evaluation. There are four steps in this method: (1) The similarity evaluation indicators are determined from three dimensions, i.e., the Basis of National Epidemic Prevention & Control, Social Resilience, and Infection Situation. (2) The data related to the indicators are collected and preprocessed. (3) The first round of screening on the preprocessed dataset is conducted through an improved collaborative filtering algorithm to calculate the preliminary similarity result from the perspective of the infection situation. (4) Finally, the K-Means model is used for the second round of screening to obtain the final similarity values. The approach will be applied to decision-making support in the context of COVID-19. Our results demonstrate that the recommendations generated by the STDSA model are more accurate and aligned better with the actual situation than those produced by pure K-means models. This study will provide new insights into preventing and controlling epidemics in regions that lack experience.

# 1  Introduction

COVID-19 is the most significant public health crisis of the 21st century(Valinejad et al., 2022). Although the variant is becoming less toxic, it still presents strong infectivity and high mortality in the elderly population(Cheung et al., 2022). Meanwhile, the virus has been observed to be more transmissible among younger individuals(Shi et al., 2021). Due to the lack of experience, resources, and other constraints, scientific responses to the outbreak become increasingly vital. Therefore, it is essential to draw experiences from other areas with similar crisis responses to support government decision-making. However, it is a big challenge to evaluate and select similar cases from massive ones.

Qualitative analysis is a widely-used method to summarize experiences at the macro level(Moury et al., 2023). Shanghai's experience is served as the basis for recommending several strategies, including applying a vulnerability analysis matrix for the targeted risk governance, empowering volunteer groups for emergency response, enacting policies and measures to prevent public health emergencies, and utilizing risk communication to facilitate uncertainty-oriented planning (L. Zhang et al., 2021). As the last country to adopt a liberal policy concerning the pandemic, China's experience in the fight against COVID-19 can serve as a reference for other countries worldwide (P. Zhang, 2021). Taizhou's approach to considering telemedicine features provides a benchmark for the management of the outbreak at home and abroad (Shao et al.). However, those studies do not cover the similarities between regions from a quantitative evaluation perspective. There is an overwhelming amount of experience to consider regarding epidemic prevention and control measures due to the numerous affected regions. Hence, it is challenging to determine the appropriate actions from a massive of cases and countries. Therefore, transfer learning, a widely-used approach based on similarity calculation, can be applied to acquire similar strategies for accurate experience learning and rapid decision-making.

Transfer learning is playing an increasingly important role in emergency management, especially in scenarios with new crises and experience shortages(Valaei Sharif et al., 2022). However, most researchers have focused on leveraging individual experience to enhance healthcare interactions (Buttery et al., 2021), nurses caring(Karimi et al., 2020), psychological caring(Sun et al., 2020), and so forth. Besides, transfer learning analyses have typically been conducted at the individual level. For example, the deep transfer learning of the Chest X-ray dataset is used to assist doctors in patient diagnosis (Jaiswal et al., 2021) or detect COVID-19 infection in the early stages (Kumar et al.). Moreover, collaborative filtering is a classical recommendation method (Luo et al., 2008). Since it was proposed by Tapestry in 1992, collaborative filtering algorithm has gradually been applied to various fields, such as movie recommendation(Behera & Nain, 2023), smart city construction(C. Zhang et al., 2022), and new coronavirus treatment plans(Yao et al., 2023). The core idea is to find precise and dependable neighbors of active users(J. Zhang et al., 2016). Data clustering is an important method in many fields, including data mining (Guizani, 2016), pattern recognition (Skomorowski, 2007), healthcare (Abbas et al., 2020), document clustering (Mahdavi & Abolhassani, 2009), image processing (Zheng et al., 2018), bioinformatics (Lam & Tsang, 2012), social networks (Xing et al., 2017), engineering (W.-L. Zhao et al., 2018), and outlier detection (Gan & Ng, 2017). K-means, a highly efficient unsupervised learning algorithm for data clustering, has been successfully developed in various real-world applications (Z. Deng et al., 2016). The K-means algorithm has been applied worldwide for cluster analysis problems (Sharma et al., 2022). However, transfer learning is rarely used from the macro dimension, such as at country or regional levels. It is challenging for governors to make decisions with inadequate prior knowledge(Green, 2021). Therefore, a macroscopic transfer learning model is required to support national or regional government decision-making. Transfer learning mainly relies on similarity calculation.

The KNN algorithm is characterized by high computational costs, substantial storage requirements, and sensitivity to imbalanced data(D. Zhao et al., 2021). Furthermore, collaborative filtering algorithms encounter the cold start problem: for new users or items, the lack of sufficient interaction data complicates the generation of effective recommendations(J. Zhang et al., 2016). Moreover, in large-scale datasets, interactions between users and items may be exceedingly sparse, affecting the accuracy of similarity calculations and recommendations. Therefore, the strengths of both algorithms are used to enhance the performance of recommendation systems or similarity computations.

In this paper, a Strategy Transfer and Decision Support Approach (STDSA) is proposed to provide technical support for national or regional governments under new crisis scenarios. Three dimensions - National Epidemic Prevention & Control, Social Resilience, and Infection Situation, are introduced to establish the similarity evaluation index system. The fundamental data are collected from different official data sources, including research papers, government reports, official news, etc. The first round of screening is performed based on the infection situation data and the approximate nearest neighbor search algorithm. An improved cooperative filtering algorithm is proposed to calculate the first round of similarity results. To obtain a more accurate result, the second round of screening is accomplished using the K-means model for clustering based on the similarity values of the first screening. The results can provide decision support for scientific epidemic prevention in a country or city with experience shortage.

## 2  Methodology

The framework and main methods of this study are introduced in this part.

*2.1 Framework of this study*

Our study is performed in the following steps, as shown in Fig.1.

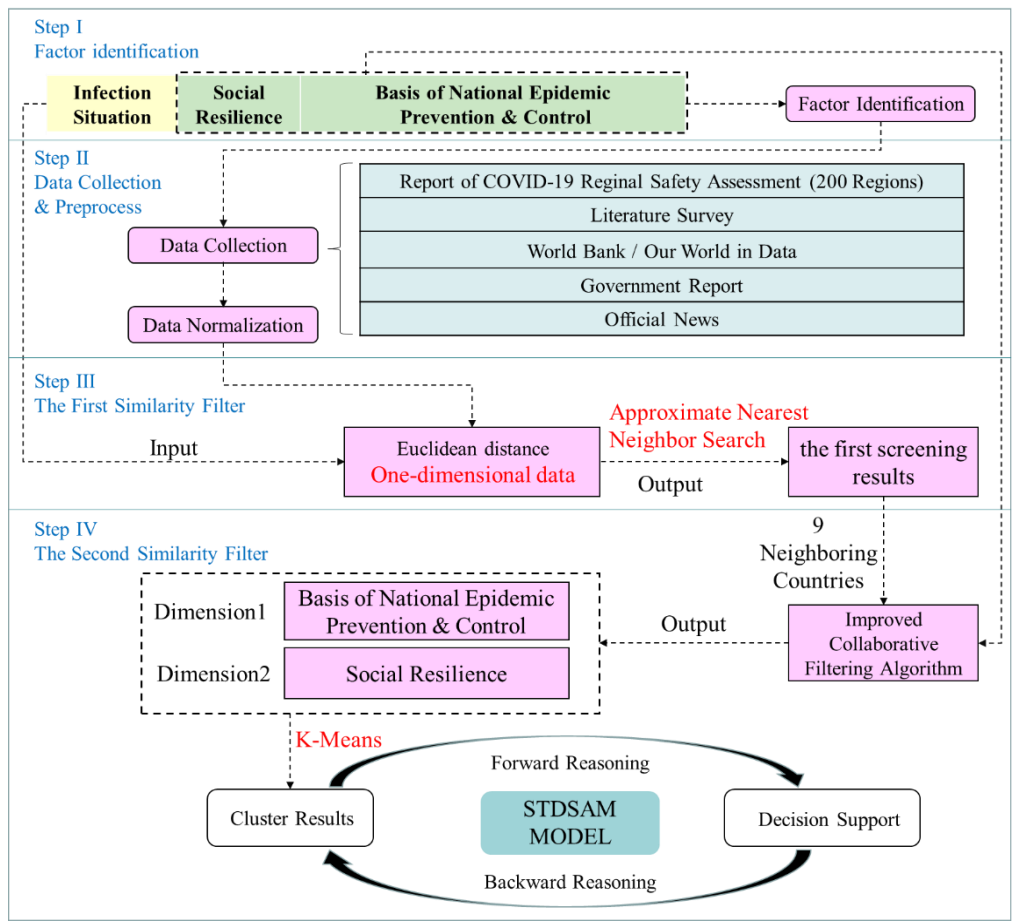

**Fig. 1.** The framework of our study

Step I: The factors are identified to evaluate the similarity from the perspective of the crisis response among different countries or regions. The relevant literature review and a United Nations report can provide a reference for our factor identification. Factors mainly include three dimensions: Basis of National Epidemic Prevention & Control, Social Resilience, and Infection Situation. The corresponding data can be collected subsequently for an in-depth study of the similarity evaluation system.

Step II: The relevant data corresponding to each evaluation factor will be collected and preprocessed for the following analysis. The data are collected from various official sources, including the COVID-19 Reginal Safety Assessment (200 Regions) report, government reports, authoritative public datasets, official news, and other sources. To eliminate the discrepancies of data, the Min-Max normalization method is employed to preprocess the collected dataset.

Step III: After the first two steps, the first similarity filter is conducted on the preprocessed dataset. It is mainly based on the Infection Situation data using the approximate nearest neighbor search algorithm. The first filtering just can remove the cases that are not similar in terms of infection situation. Therefore, a second filtering algorithm is required based on the first screening result.

Step IV: An improved collaborative filtering algorithm combined K-Means model is proposed for the secondary similarity filter. The similarity value ranges from -1 to 1. To progress to the subsequent steps, the accurate similarity value should be calculated to generate a new specific dataset. Following the first round of screening, the effects of Infection data are nullified. Therefore, the improved collaborative filtering algorithm is used to quantify the similarity among 9 neighboring countries (the result after the first screening) based on the data of the Basis of National Epidemic Prevention & Control and Social Resilience. Then the K-means model is used to calculate the resemblance value. Subsequently, a more precise similarity categorization can be obtained through the secondary screening.

When a specific scenario is described quantitatively based on the previous state assignment method, the scenario can be fed into our model. Then a transfer strategy can be recommended to restrain or impede the spread of a viral infection in the target country or region with experience shortage.

*2.2 Collaborative Filtering Algorithm*

The user-based collaborative filtering algorithm is adopted in this paper. When calculating the users' similarity, there are primarily three types of similarity measures between users: the standard cosine similarity algorithm, the adjusted cosine similarity algorithm, and the Pearson similarity algorithm(Chen et al., 2016). Among them, the Pearson similarity algorithm is the most widely-used measure to exhibit the linear correlation., which will be used in our study (Di Lena & Margara, 2010). The algorithm is shown in Equation (1).

$$\text{sim}(i,j) = \frac{\sum_{x \in I_{ij}} (R_{i,x} - \bar{R}_i)(R_{j,x} - \bar{R}_j)}{\sqrt{\sum_{x \in I_{ij}} (R_{i,x} - \bar{R}_i)^2} \sqrt{\sum_{x \in I_{ij}} (R_{j,x} - \bar{R}_j)^2}} \quad (1)$$

where $R_{i,x}$ is the rating of user $i$ on item $x$. $\bar{R}_i$ is the average of users' ratings. $I_{i,j}$ indicates the items that user $i$ and $j$ co-evaluated. $I_i$ is item set that user $i$ rated. $I_j$ is item set that user $j$ rated.

*2.3 K-Means Algorithm*

Since there are not enough training datasets or predefined labels, K-means, as an unsupervised classification algorithm, can be used to conclude the small amount of data (Fahim, 2021). The k-means objective function is shown in Equation (2).

$$J(Z,A) = \sum_{m=1}^{n} \sum_{k=1}^{c} z_{mk} \|R_m - a_k\|^2 \quad (2)$$

where $\mathbf{R}' = \{R_1, \ldots, R_n\}$ is a dataset in a d-dimensional Euclidean space $D^d$. $n$ is the number of samples. $A = \{a_1, \ldots, a_c\}$ is the $c$ cluster centers. And each element of A is the central point of each class (Sinaga & Yang, 2020). $Z = z_{mk}$ is a binary variable, which indicates whether the data point $R_m$ belongs to $k$-th cluster($m = 1,2,\ldots,n;\ k = 1,\cdots,c$).

The k-means algorithm is iterated for minimizing the objective function $J(Z,A)$ with updating equations for cluster centers and memberships, respectively.

$$a_k = \frac{\sum_{i=1}^{n} z_{mk} R_{mk}}{\sum_{i=1}^{n} z_{mk}} \quad (3)$$

and

$$z_{mk} = \begin{cases} 1 & \text{if} \|R_m - a_k\|^2 = \min_{1 \leq k \leq c} \|R_m - a_k\|^2 \\ 0, & \text{otherwise.} \end{cases} \qquad (4)$$

where $\|R_m - a_k\|$ is the Euclidean distance between the data point $R_m$ and the cluster center $a_k$. And for unsupervised evaluation indexes, Sum of Squares for Error(SSE) is directly used to evaluate and analyze the clustering results. It could be defined as follows(Ran et al., 2021).

$$SSE = \sum_{k=1}^{C} \sum_{q=1}^{Q} \left(R_{kq} - \overline{R_k}\right)^2 \qquad (5)$$

where $c$ represents the number of clusters, $Q$ represents the number of data points in clustering, and $\left(R_{kq} - \overline{R_k}\right)^2$ represents the Sum of Squares for Error for each data point. Since the number of clusters could be set manually, Hussain and Haris proposed that it would be the number when the descent suddenly slows down.

# 3 Model Building

*3.1 Factor Identification*

Factor identification primarily relies on the report of COVID-19 Reginal Safety Assessment (200 Regions), published by Deep Knowledge Group, a consortium of commercial and non-profit organization (Hezer, Gelmez, & Ozceylan, 2021). For safety assessment in the context of COVID-19, six evaluation indicators have been proposed, namely Quarantine Efficiency, Government Efficiency, Monitoring and Detection, Healthcare Readiness, Regional Resiliency, and Emergency Preparedness. Detailed components of each indicator are depicted in Figure 2. For instance, the indicators in Quarantine Efficiency include Scale of Quarantine, Quarantine Timeline, Criminal Penalties for Violating Quarantine, Economic Support for Quarantined Citizens, Economic Supply Chain Freezing, and Travel Restrictions. These indicators are mainly related to the infrastructure for first aid medical treatment and government management.

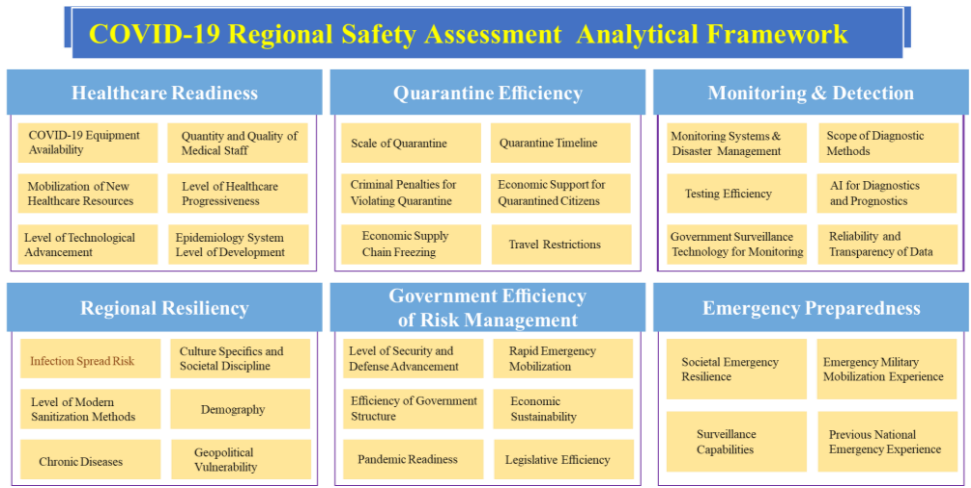

**Fig. 2.** The detailed integrant of the report

The specific index system and the sources of data collection in our study are shown in Table 1. The infrastructure system plays a critical role in assessing the risks of infectious disease transmission(Clamp & Crees, 2020). Rigorous government measures can significantly reduce the impact of epidemics (Q. Deng et al., 2023). However, the social resilience also plays a critical role in COVID-19 related model (Q. Deng et al., 2023). And as for a given region, it is observed that the Disease-related information also plays an important part in pandemic prevention (Backholer et al., 2021). In STDSA, three dimensions are considered to evaluate similarity across nations or regions, listed as National Epidemic Prevention & Control, Social Resilience, and Infection Situation. For a region with missing experience, these dimensions enable informed decision-making in response to threats.

The Government risk management efficiency, Emergency preparedness, Quality and Accessibility of Care Index, and Monitoring and diagnosis criteria are used to assess the Basis of the National Epidemic Prevention dimension. The Social Resilience dimension can be evaluated by the indicators, such as the percentage of the population under a certain age (Gao et al., 2020), educational attainment(Alrouh et al., 2022), population density (Hu et al., 2013), and mass living level (Furuse, 2019). Lastly, the Infection Situation is measured using the proportion of infection, which is the number of confirmed cases divided by the total population.

**Table 1** STDSA Evaluation System

| Dimension | Category | Elements | Data Sources |
|---|---|---|---|
| $\alpha$ | Basis of National Epidemic Prevention & Control | Government Risk Management Efficiency<br>Emergency Preparedness<br>Quality and Accessibility of Care Index<br>Monitoring and Diagnosis | Report of COVID-19 Reginal Safety Assessment |
| $\beta$ | Social Resilience | Low Age Distribution<br>Education Level<br>Population Density<br>Mass Living Level | World Bank (https://data.worldbank.org/)<br>Our World in Data (https://ourworldindata.org/) |
| $\gamma$ | Infection Situation | Number of Infections Divided by the Total Number of People | https://github.com/eAzure/COVID-19-Data |

*3.2 Data Collection and Preprocessing*

*3.2.1 Data Collection*

The data related to various indicators are acquired from several reliable sources, as shown in Table 1. The Basis of National Epidemic Prevention dimension is predominantly sourced from the COVID-19 Regional Safety Assessment Report (200 Regions). Meanwhile, the World Bank (https://data.worldbank.org/) and Our World in Data (https://ourworldindata.org/) serve as the primary sources for the other two dimensions. Furthermore, government reports and official news are also utilized to verify the accuracy of the data. The collected data can be used to construct a dataset for further analysis, with 10 samples in the dataset displayed in Table 2. The complete dataset is available on the GitHub website (https://github.com/Crystalxy123/Data-for-STDSAM--).

**Table 2** the Dataset of STDSA

| Region | Infection (I) | Government Risk Management (G) | Emergency Preparedness (P) | Quality and Accessibility of Care (Q) | Education Level (E) | Young Distribution (Y) | Population Density (P) | Mass living Level (M1) | Monitoring and Diagnosis (M2) |
|---|---|---|---|---|---|---|---|---|---|
| Germany | 0.21 | 194.00 | 79.00 | 92.00 | 24.00 | 14.00 | 240.40 | 4.60 | 102.00 |
| France | 0.35 | 113.00 | 78.00 | 92.00 | 22.95 | 17.65 | 119.20 | 3.90 | 82.00 |
| China | 0.00 | 172.00 | 139.00 | 78.00 | 8.00 | 18.00 | 148.00 | 1.14 | 132.00 |
| Japan | 0.05 | 184.00 | 91.00 | 94.00 | 30.59 | 12.45 | 347.00 | 4.00 | 142.00 |
| India | 0.03 | 131.00 | 99.00 | 41.00 | 8.50 | 26.16 | 464.10 | 0.19 | 80.00 |
| America | 0.25 | 100.00 | 103.00 | 89.00 | 39.00 | 39.00 | 36.20 | 6.40 | 86.00 |
| Canada | 0.09 | 172.00 | 100.00 | 94.00 | 45.00 | 16.00 | 4.20 | 4.32 | 133.00 |
| Australia | 0.14 | 181.00 | 118.00 | 96.00 | 34.30 | 19.29 | 3.30 | 5.18 | 116.00 |
| Israel | 0.41 | 191.00 | 113.00 | 85.00 | 35.01 | 27.83 | 400.00 | 4.36 | 143.00 |
| Thailand | 0.05 | 144.00 | 60.00 | 69.00 | 21.11 | 14.68 | 136.60 | 0.72 | 95.00 |

*3.2.2 Data Statistical Description*

Box line diagram is a useful visualizing tool for anomalies and data distribution(Bonciani et al., 2004). It is utilized to analyze the statistical characteristics and data quality of our dataset. Fig. 3 intuitively shows that there are no outliers within the Infection and Quality and Accessibility of Care Index categories. However, outliers have been identified in the Population density and Monitoring and diagnosis groups, which are significantly abnormal. Outliers are also observed in the remaining index and must be properly processed to eliminate their influence. Interestingly, the median line for the Government Risk Management Efficiency, Quality and Accessibility of Care Index, and Education Level categories fall within the middle of the box, suggesting a normal distribution of data. On the contrary, the median line for the Population Density category is higher on the box, indicating a right-skewed distribution.

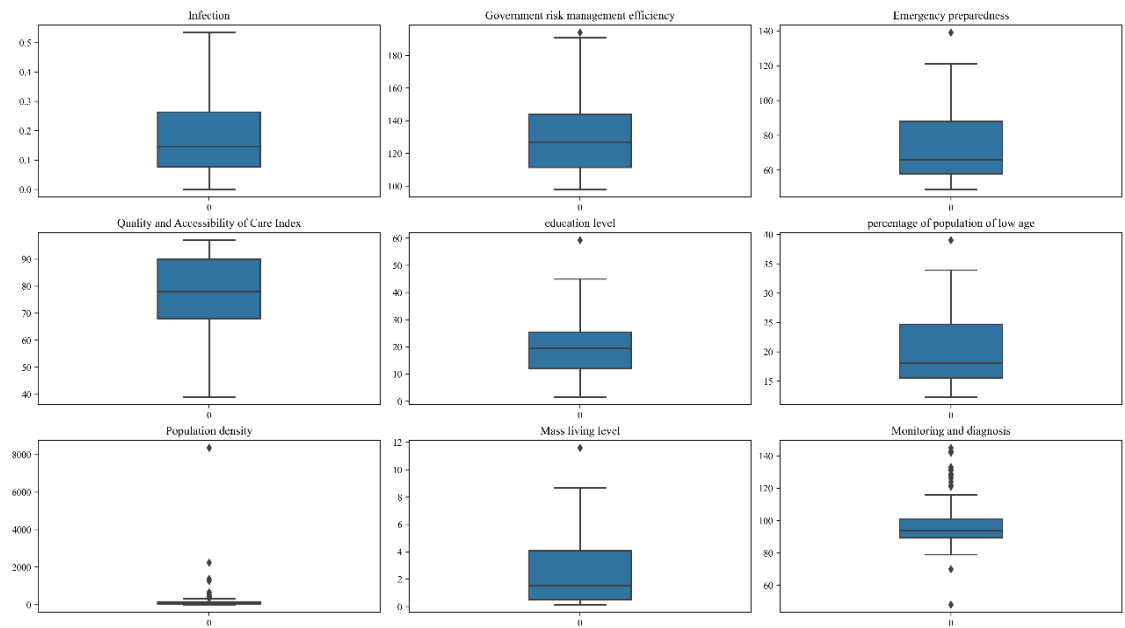

**Fig. 3.** Distribution of indicators

To analyze the correlations between indicators, Pearson Correlation Coefficient (PCC) is utilized in this paper. This measure ranges from -1 to +1(Leonenko et al., 2013). It signifies there is a positive correlation between two factors when the value of PCC exceeds 0, while a negative correlation exists when PCC is smaller than 0 (Armstrong, 2019). The strength of the correlation can be determined by the absolute value of PCC. A greater absolute value indicates a stronger relationship (Britten et al., 2017). The Person correlation between variables is shown in Fig.4. Each variable with abbreviations is depicted in Table 2. Different colors in the heat map show different values. Dark yellow indicates a higher similarity value, while blue denotes lower values.

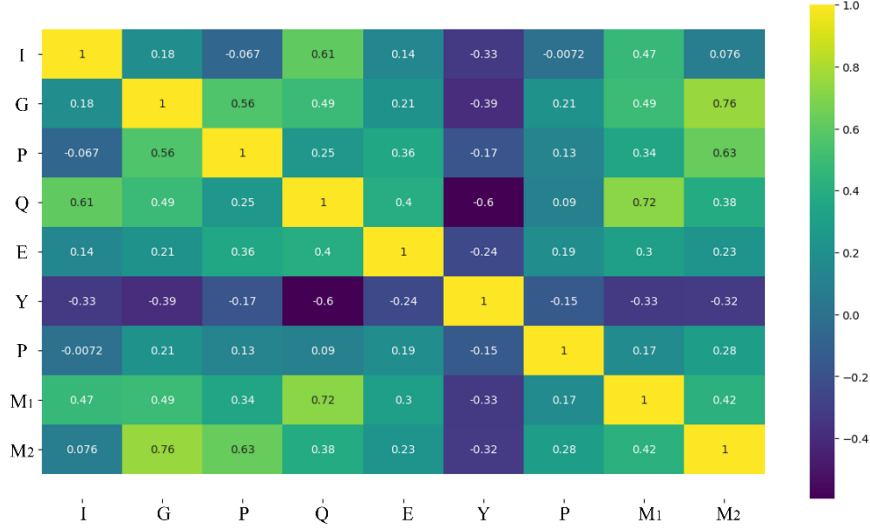

**Fig. 4.** Heat map of Person coefficients between variables

It is observed that there are two groups with high positive correlations: Quality and Accessibility of Care Index ⇔ Infection/Mass Living Level and Monitoring and Diagnosis ⇔ Government Risk Management Efficiency/Emergency Preparedness. Conversely, a high negative correlation can be seen in the Quality and Accessibility of Care Index ⇔ Percentage of Population of Low Age group.

There are some low positive correlations, which are Emergency Preparedness ⇔ Quality and Accessibility of Care Index/Education Level/Mass Living Level, Monitoring and Diagnosis ⇔ Quality and Accessibility of Care Index/Education Level/Population Density, Government Risk Management Efficiency ⇔ Education Level/Population Density, and Education Level ⇔ Mass Living Level. Meanwhile, there are low negative correlations between Percentage of Population of Low Age ⇔ Infection/Government Risk Management Efficiency/Education Level and Percentage of Population of Low Age ⇔ Mass Living Level/Monitoring and Diagnosis. The other indicator group combinations demonstrate essentially uncorrelated results.

*3.2.3 Data Preprocessing*

Data preprocessing is an essential step in model design. And normalization is a frequently used technique to handle out-of-bounds attributes (Z. Zhang et al., 2014). In this study, the Min-Max normalization method is used to scale the inputs using Gaussian distribution (Preetha & Mallika, 2021). This normalization technique is implemented in two main steps.

Step 1: Following Eq. (5), the maximum and minimum bounds are calculated using STDSA raw dataset. The output of the first step can be the global maximum and minimum values.

Step 2: Since the global maximum and minimum values are determined, STDSA datasets can be normalized using the standard Min-Max normalization.

$$\widehat{R_{I_i}} = \frac{R_{I_i} - \min(R_I)}{\max(R_I) - \min(R_I)} \tag{6}$$

where $\widehat{R_{I_i}}$ is the normalized value for $I_i$ indicators' value of country $i (R_{I_i})$. Maximum value is represented as $\max(R_I)$, and minimum value is $\min(R_I)$. Then, similarity calculation stage is given with these output values as its inputs. The normalized dataset is shown in Table 3.

Table 3 The normalized dataset of STDSA

| Region | Infection (I) | Government risk management (G) | Emergency Preparedness (P) | Quality and Accessibility of Care (Q) | Education level (E) | Young Distribution (Y) | Population density (P) | Mass living level (M1) | Monitoring and diagnosis (M2) |
|---|---|---|---|---|---|---|---|---|---|
| Germany | 0.39 | 1.00 | 0.33 | 0.91 | 0.39 | 0.06 | 0.03 | 0.39 | 0.56 |
| France | 0.65 | 0.16 | 0.32 | 0.91 | 0.37 | 0.20 | 0.01 | 0.33 | 0.35 |
| China | 0.00 | 0.77 | 1.00 | 0.67 | 0.11 | 0.21 | 0.02 | 0.09 | 0.87 |
| Japan | 0.09 | 0.90 | 0.47 | 0.95 | 0.50 | 0.01 | 0.04 | 0.34 | 0.97 |
| India | 0.06 | 0.34 | 0.56 | 0.03 | 0.12 | 0.52 | 0.06 | 0.00 | 0.33 |
| America | 0.46 | 0.02 | 0.60 | 0.86 | 0.65 | 1.00 | 0.00 | 0.55 | 0.39 |
| Canada | 0.17 | 0.77 | 0.57 | 0.95 | 0.75 | 0.14 | 0.00 | 0.37 | 0.88 |
| Australia | 0.26 | 0.86 | 0.77 | 0.98 | 0.57 | 0.26 | 0.00 | 0.44 | 0.70 |
| Israel | 0.75 | 0.97 | 0.71 | 0.79 | 0.58 | 0.58 | 0.05 | 0.37 | 0.98 |
| Thailand | 0.09 | 0.48 | 0.12 | 0.52 | 0.34 | 0.09 | 0.02 | 0.05 | 0.48 |

*3.3 First Similarity Filter*

A first similarity filter is used to reduce the computational workload of subsequent tasks. Because it can remove the cases that are not similar in terms of infection situations. The first filter employs the approximate nearest neighbor search algorithm, which is a distance-based method for classification tasks (S. Zhang et al., 2017). It is regarded as one of the top 10 data-mining algorithms (X. Wu et al., 2008). The main task of the approximate nearest neighbor search classification is to predict the labels of test data points by considering all the training data points. It is well known that the classification method has at least two open issues to be addressed (S. Zhang, 2010), which are the similarity measurement between two data points and the selection of the k value. For the purposes of our study, the algorithm's core idea is to find the nearest neighbors of a query in the training data.

As illustrated by Fig.5, the normalized dataset of Infection Situation dimension is supplied as the input of the first similarity filter. Subsequently, the distance between our target region and other regions via multi-search is computed. Finally, the parameter p is set in first similarity filter, which is the number of nearest neighboring regions. In our paper, after careful consideration of both computational workload and result accuracy, p(neighbor_number) is set as 8. The partial results are listed as Table 4.

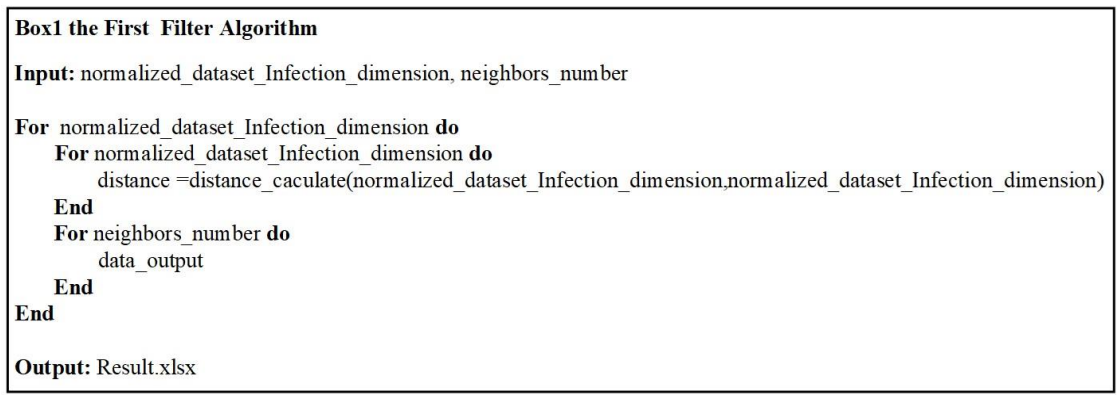

**Fig. 5.** The First Similarity Filter Algorithm

**Table 4 The results of first similarity filter**

| Country | Neighbor$_1$ | Neighbor$_2$ | Neighbor$_3$ | Neighbor$_4$ | Neighbor$_5$ | Neighbor$_6$ | Neighbor$_7$ | Neigbor$_8$ |
|---|---|---|---|---|---|---|---|---|
| Peru | Ukraine | Belarus | Bosnia and Herzegovina | Albania | Malaysia | Russia | Paraguay | Qatar |
| Algeria | Egypt | Cambodia | Myanmar | Taiwan | Chinese Mainland | The People's Republic of Bangladesh | Saudi Arabia | Indonesia |
| Panama | Chile | Turkey | Hungary | Jordan | Singapore | Bulgaria | Lebanon | Poland |
| Albania | Bosnia and Herzegovina | Belarus | Paraguay | Canada | The United Arab Emirates | Tunisia | Peru | Iran |
| Egypt | Algeria | Cambodia | Taiwan | Chinese Mainland | Myanmar | The People's Republic of Bangladesh | Saudi Arabia | Indonesia |
| Belgium | Luxembourg | Cyprus | Bahrain | Britain | Slovakia | The Czech Republic | Ireland | Portugal |

*3.4 Second Similarity Filter*

The first similarity filter can help to remove the cases that are not similar in terms of infection situation. However, three dimensions including Basis of National Epidemic Prevention & Control, Social Resilience, and Infection Situation are needed in the similarity calculation. Therefore, a second round of similarity calculation is required to filter the cases that are not similar with the target country in terms of Basis of National Epidemic Prevention & Control and Social Resilience. In this process, the improved collaborative filtering algorithm is used to quantify the similarity among 9 neighboring countries (the first screening result) based on the data related to these two dimensions. Then the K-Means is conducted to generate a specific set of similar regions further. As an unsupervised learning method, K-Means is one of the most widely-used clustering algorithms. It can be used to group N data points into c cluster by minimizing the sum of squared distances between each point and its nearest cluster center. After the secondary screening, it can support decision-making for our targeted region to take epidemic control measures in an experience shortage scenario. Specifically, the given region can generate strategies based on the measures executed by the similar regions to prevent or slow down the propagation of a viral infection.

The collaborative filtering recommendation algorithm is a widely-used method for similarity calculation (Fu et al., 2019). However, due to the multidimensionality of our indicators, the traditional collaborative filtering algorithm is not fit for our problem. An improved collaborative filtering algorithm is introduced. The similarity between different countries is calculated as

$$sim_d(i,j) = \frac{\sum_{x \in I_{i,j}}^{y=d}(R_{i,x,y} - \overline{R_{i,y}})(R_{j,x,y} - \overline{R_{j,y}})}{\sqrt{\sum_{x \in I_{i,j}}^{y=d}(R_{i,x,y} - \overline{R_{i,y}})^2}\sqrt{\sum_{x \in I_{i,j}}^{y=d}(R_{j,x,y} - \overline{R_{j,y}})^2}} \quad (7)$$

where $sim_d(i,j)$ is the similarity between country $i$ and country $j$ in $d$ dimension($d$=1,2). And $I_{i,j}$ means indicators shared by country $i$ and country $j$. As the ones mentioned in Section 2.3, $R_{i,x,y}$ named $y$ dimension's $x$ indicator of country $i$. Furthermore, $\overline{R_{i,y}}$ represents the average value of each indicator in dimension $y$.

Fig.6 illustrates the similarity values of Dimension $\alpha$ and Dimension $\beta$. A heat map is employed for intuitive visualization, with yellow indicating higher values and blue indicating lower values. The similarity values range from -1 to 1. The results of the first filtering stage display greater similarity within Dimension $\alpha$ compared to Dimension $\beta$. It implies that the Basis of National Epidemic Prevention & Control dimension is more similar than the Social Resilience dimension. This will be further explored in the succeeding section via the second similarity filter.

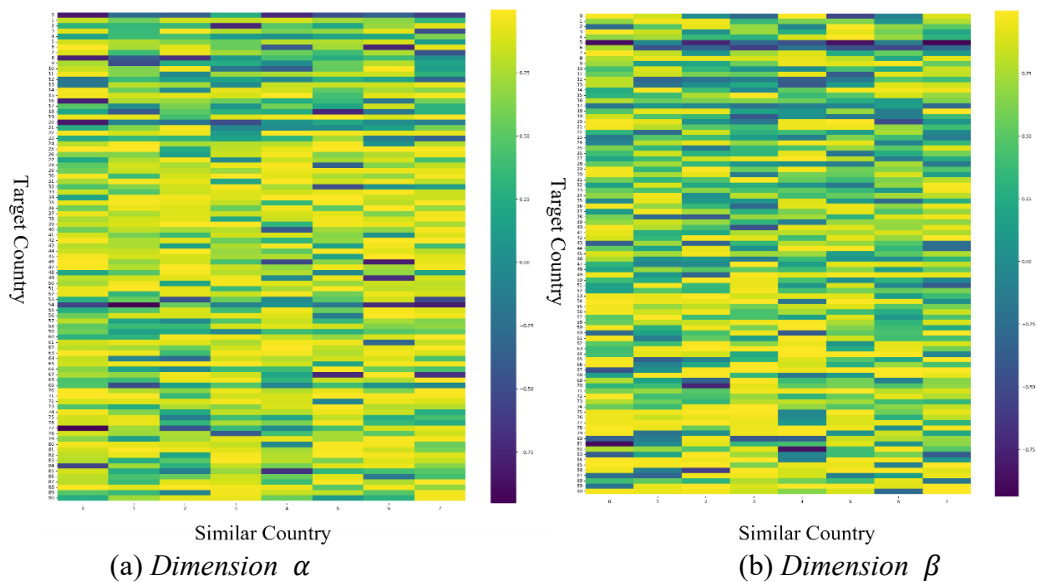

(a) *Dimension $\alpha$*  (b) *Dimension $\beta$*

**Fig. 6.** Similarity of Dimension $\alpha$ & Dimension $\beta$

# 4 Case Study and Result Analysis

*4.1 Case Study*

Two regions are selected for our numerical experiments: Sweden and Mainland China. The collected data are preprocessed using Min-Max normalization method, as described in Section 3.2.3. Subsequently, the first similarity filter algorithm of STDSA is used. Based on the results presented in Table 4, the similar regions to Sweden include the United States, Uruguay, Norway, Greece, Spain, Croatia, Ireland, and Italy. Mainland China is surrounded by eight neighboring countries, which are Taiwan, Egypt, Algeria, Cambodia, Myanmar, Bangladesh, Saudi Arabia, and Indonesia.

As discussed in Section 2.2, an improved collaborative filtering algorithm and K-means are utilized to obtain the specific similarity value based on the first similarity filter results. As shown in Table 5, it can be found that Spain, Ireland, and Italy are similar to Sweden in both measured dimensions (more than 0.88). A greater similarity with the United States and Croatia is observed in the National Base dimension (about 0.80), while the similarity of Mass Base is lower than 0.50 for both. Interestingly, For Mainland China, there are three regions with a low similarity rate, which are Taiwan, Egypt, and Algeria. Additionally, the qualification results are not as favorable as those of the Sweden region. Nevertheless, Myanmar and Bangladesh can still be identified as the desired regions.

**Table 5** The results of similarity calculation

| Region | Similar Regions | National Base | Mass Base |
| --- | --- | --- | --- |
| Sweden | United States | 0.80 | 0.50 |
| | Uruguay | 0.73 | 0.39 |
| | Norway | 0.55 | 0.98 |
| | Greece | 0.85 | 0.66 |
| | Spain | 1.00 | 0.94 |
| | Croatia | 0.81 | 0.08 |
| | Ireland | 0.92 | 0.92 |
| | Italy | 0.99 | 0.88 |
| Mainland China | Taiwan | -0.77 | 0.77 |
| | Egypt | -0.39 | 0.94 |
| | Algeria | -0.66 | 0.94 |
| | Cambodia | 0.10 | 0.87 |
| | Myanmar | 0.41 | 0.93 |
| | Bangladesh | 0.27 | 0.80 |
| | Saudi Arabia | 0.39 | 0.21 |
| | Indonesia | 0.02 | 0.91 |

*Note: those with red color have an overwhelmingly low similarity rate*

Subsequently, K-means algorithm is employed as the second filter of STDSA. As shown in Fig.7, it is evident that the SSE decreases with the number of clusters increases. Upon reaching four clusters, however, the decrease in SSE becomes insignificant. Therefore, four is the relative best number of classifications.

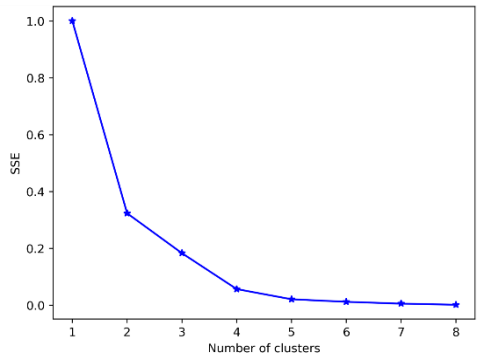 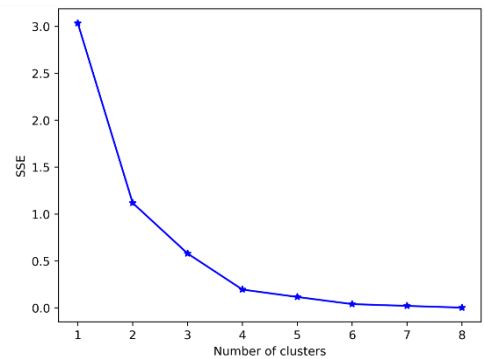

(a) Sweden  (b) Mainland China

**Fig. 7.** Elbow for K-means Cluster

When the cluster number, 4, is set in STDSA, the specific results are displayed in Fig.8. The different color indicates different cluster. Specifically, there are four colors in Fig.8, which is consistent with the number of clusters set above.

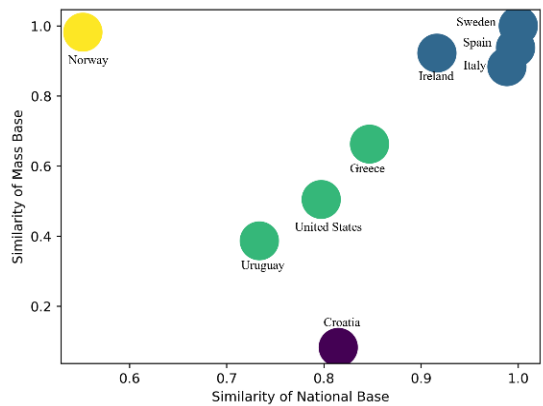 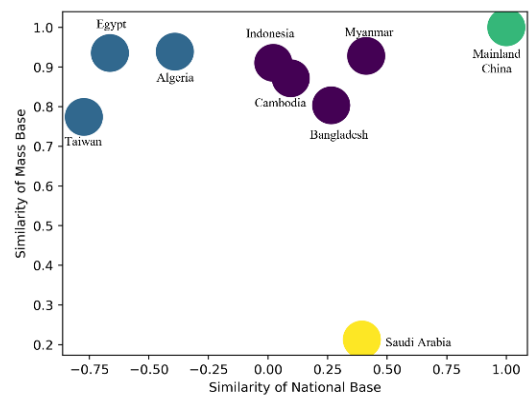

(a) Sweden  (b) Mainland China

**Fig. 8.** Results of the Second Filter

(a)The second filter result of Sweden

Sweden, Spain, Ireland, and Italy are displayed with the same yellow color in Fig.8, which represents that Sweden exhibits a high level of similarity with Spain, Ireland, and Italy. The reasons may be as follows.

Sweden and Italy are two different countries, which do not share a border geographically. Nevertheless, both share a great enthusiasm for football. Although Sweden and Italy are geographically distant countries without a common border, they share similar interests in various aspects, including football, meatballs, coffee, and design. In addition, both countries have shown great interests in potential ecological tax reforms in the 1990s (Gren et al., 2003). Another similarity between the two countries is their comparable Rare Earth Element (REE) concentrations (Sadeghi et al., 2015).

As for Ireland and Sweden, two countries speak the incredibly similar languages. According to a Statista survey on hobbies and interests, around 10% of respondents agree that cooking is a great way to spend their free time (https://www.statista.com/statistics/). The number of Irish people residing in Sweden has also increased significantly, with approximately 2,500 Irish residents recorded. Furthermore, both countries rank in the top ten globally for providing the highest quality of life to their residents(Pedro et al., 2020).

It is worth noting that Spain is approximately 2,675 kilometers from Sweden. In 2014, similar impacts of emissions from shipping on SO2 concentrations were observed in both Sweden and Spain (Viana et al., 2014), indicating similar eco-friendly philosophies. Moreover, both countries prioritize land management with different approaches, such as organic farming in Spain and wetland construction in Sweden (Keesstra et al., 2018). Another noteworthy point is that both Spain and Ireland maintained their neutrality during the Second World War.

(b)The second filter result of Mainland China

It reveals that there is no region similar to Mainland China. However, this unexpected result can be explained by several factors. It has confirmed that the stricter government measures can achieve better epidemic prevention and control effects (Q. Deng et al., 2023). As of December 2022, Mainland China is the only country that has maintained a consistent approach towards outbreak prevention and control (J. Wu et al., 2021), while other regions chose to forgo containment. Further, Mainland China is one of the few socialist countries in the world, characterized by a socialist system and socialist traits (Yang et al., 2010). Moreover, as the world's most populous nation with a high population density, Mainland China has distinctive features that ensure the uniqueness of the country.

*4.2 Model Comparison*

To verify the practicality of the proposed STDSA, the fully K-means method is introduced to compare with our STDSA. The main idea of the fully K-means method is to realize the cluster analysis only using K-means. The specific results are shown in Fig. 9. As illustrated in Fig. 9. (a), it is evident that SSE decreases as the number of clusters increases. However, upon reaching five clusters, the decrease in SSE becomes insignificant. Therefore, five can be considered as the optimal number of classifications. Additionally, as shown in Fig. 9. (b), the clustering results of K-means are found to be relatively broad. Furthermore, it can be observed that the classification division is not uniform. For instance, Category 1 includes 28 countries, while Category 5 only contains one country. Specifically, the neighboring countries of Mainland China and Sweden are presented in the Table 6.

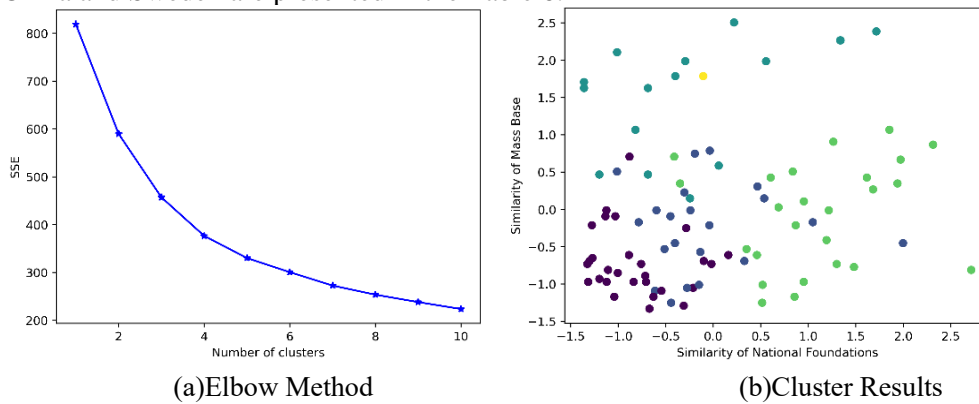

(a)Elbow Method      (b)Cluster Results

**Fig. 9.** Results of the K-means

**Table 6** The distribution of K-means Results

| Category | number |
|---|---|
| Category1 | 28 |
| Category2 | 21 |
| Category3 | 16 |
| Category4 | 25 |
| Category5 | 1 |

According to the k-means clustering results, there are 24 countries similar to Sweden, and all the countries recommended by our STDSA model occur in this result. Additionally, it is imprecise to claim that Sweden, Qatar, and Montenegro have similar COVID-19 prevention measures. Despite certain levels of trade and investment cooperation, the overall connection and exchange between Sweden and Qatar is relatively minimal due to factors such as geographical distance and cultural disparities. Similarly, while Sweden and Montenegro maintain some form of cooperation in areas like trade and tourism, their relationship is weak. Furthermore, there are 15 countries, including Canada, Australia, Germany, etc, similar to Mainland China, as shown in Table 7. However, it is evident that there is a low level of mutual learning and borrowing between China and Canada for epidemic prevention measures.

Table 7 The contrast and intrinsic dim of cases

| Index | Similar Regions of Sweden | Similar Regions of Mainland China |
|---|---|---|
| 1 | United Kingdom | Germany |
| 2 | Denmark | Switzerland |
| 3 | Netherlands | Israel |
| 4 | Iceland | Japan |
| 5 | Finland | Austria |
| 6 | Luxembourg | Australia |
| 7 | Qatar | New Zealand |
| 8 | Lithuania | South Korea |
| 9 | Latvia | United Arab Emirates |
| 10 | Slovenia | Canada |
| 11 | Estonia | Norway |
| 12 | Croatia | Taiwan |
| 13 | Ireland | Saudi Arabia |
| 14 | Cyprus | Hungary |
| 15 | Montenegro | Kuwait |
| 16 | Czech Republic | |
| 17 | Spain | |
| 18 | Portugal | |
| 19 | Italy | |
| 20 | Slovakia | ------ |
| 21 | United States | |
| 22 | France | |
| 23 | Belgium | |
| 24 | Andorra | |

# The blue color represents the results that are consistent with STDSA. And the red color indicates that the country is less similar to the target country in terms of epidemic prevention and control.

Based on our comparative analysis of the STDSA results and pure K-means results, it is evident that the output generated by STDSA provides a more precise range of recommendations that aligns better with the actual situation. Therefore, the STDSA can be considered as an effective tool for generating accurate suggestions for countries to learn from each other's pandemic response measures. The incorporation of additional data sources and analytical techniques in future studies can further improve the model's performance and generate even more precise and reliable results.

*4.3 Result Analysis and Evaluation*

The final K-means cluster results can be evaluated by two factors, which are contrast and intrinsic dim (Fränti & Sieranoja, 2018). Contrast is defined as the relative difference in the distance to its nearest and furthest neighbor. Intrinsic is a measurement to estimate the true dimensionality of the data. And it is calculated as the (squared) average distance among all points divided by the variance of the distances (Buchsbaum & Snoeyink, 2001). The contrast and intrinsic dim are presented in Table 8.

Table 8 The contrast and intrinsic dim of cases

| Region | Contrast | Intrinsic dim |
|---|---|---|
| Sweden | 15.20 | 2.17 |
| Mainland China | 2.02 | 2.80 |

The results of contrast and intrinsic dim demonstrate that the contrast of STDSA classification result for Sweden is 15.20, while Mainland China has a significantly lower contrast, approximately 2.00. It suggests that the STDSA results of Mainland China are more uniform compared to Sweden. Additionally, both regions have similar values in the intrinsic dimension, which indicates that the complexity of the results for both regions is similar.

In conclusion, the response strategies of Sweden can provide some references for Spain, Ireland, and Italy. However, Mainland China's distinct characteristics make it a poor reference point for others because of its uniqueness. Nevertheless, Mainland China can still offer valuable insights on national-level epidemic prevention and control. It is worth noting that the measures implemented by the target region can significantly influence the outcomes of STDSA classification.

## 5 Discussion and Conclusion

As the shortage of experienced personnel can severely affect a region's decision-making, the STDSA is proposed as a decision support system. Two filters are utilized to concurrently perform data clustering. Following the evaluation index identification, the dataset is collected and preprocessed using Min-Max normalization method. Afterward, an approximate nearest neighbor search is used to conduct a preliminary screening on Infection Situation data, followed by an improved collaborative filtering recommendation algorithm to qualify the similarity. Finally, the K-Means algorithm is used to carry out a second screening. The case study of STDSA is conducted by estimating two samples in an experience shortage scenario. The reasons behind the STDSA results are analyzed. In conclusion, the initial results show that STDSA can evaluate the similarity between distinct regions from an epidemic prevention point of view. And it is evident that the output generated by STDSA model provides a more precise range of recommendations that aligns better with the actual situation compared with pure K-means method. Therefore, the STDSA model can be considered an effective tool for generating accurate suggestions for countries to learn from each other's pandemic response measures. And the measures taken by the target region will influence the results of STDSA classification.

By fusing historical case data with expert knowledge, the results generated by STDSA can provide useful information and references for decision-making. However, there are several limitations of STDSA. For example, national-level epidemic-spreading cases are needed to refine the identified factors. A larger sample size will improve the model's performance. It is crucial to investigate how to select an appropriate value for the parameter "p" required by the STDSA algorithm. Currently, "p" is assumed as an input parameter. Due to the unsupervised nature of STDSA, evaluation becomes difficult as there are no labels for verification and comparison. The reasonableness of the results can only be analyzed in the light of practical experience. Furthermore, since each region has unique characteristics in the temporal dimension, it is important to consider local characteristics to develop the most appropriate strategy.

Besides, STDSA is the similarity evaluation system considering the characteristics of viruses with transmissibility such as population density. The model is only applicable to infectious diseases with transmissibility.

## Statements and Declarations

**Funding.** This work is supported by National Science Foundation of China (Grant Nos. 72004113, 72274123, 72174099, 71904121, 71904193) and High-tech Discipline Construction Fundings for Universities in Beijing (Safety Science and Engineering). In addition, thanks for the support of the University of Science and Technology Beijing undergraduate Innovation and Entrepreneurship Training Program.

**Competing Interests.** The authors have no relevant financial or non-financial interests to disclose.

**Author Contributions.** All authors contributed to the study conception and design. Material preparation, data collection and analysis were performed by Xingyu Xiao, Peng Chen, Xue Cao and Kai Liu. The first draft of the manuscript was written by Xingyu Xiao and all authors commented on previous versions of the manuscript. All authors read and approved the final manuscript.